\documentclass[wcp]{jmlr}

\usepackage{amsfonts}
\usepackage{mathtools}
\usepackage{booktabs}
\usepackage{multirow}
\usepackage{makecell}
\usepackage{adjustbox}
\usepackage{textcomp}
\usepackage{xcolor}
\usepackage{tikz}
\usetikzlibrary{arrows.meta,positioning,shapes.geometric,fit,calc,patterns}
\usepackage{pgfplots}
\pgfplotsset{compat=1.18}

\usepackage{lineno}
\makeatletter
\let\Ginclude@graphics\@org@Ginclude@graphics
\makeatother

\jmlryear{2026}
\jmlrworkshop{ACML 2026}

\definecolor{mplblue}{HTML}{1F77B4}
\definecolor{mplorange}{HTML}{FF7F0E}
\definecolor{mplgreen}{HTML}{2CA02C}
\definecolor{mplred}{HTML}{D62728}
\definecolor{mplpurple}{HTML}{9467BD}
\definecolor{mplbrown}{HTML}{8C564B}
\definecolor{mplpink}{HTML}{E377C2}

\title[Embedding Semantic Entropy]{Embedding Semantic Entropy: NLI-Free Semantic Uncertainty Estimation with Geometry-Aware Clustering}

\author{%
  \Name{Akshiya Raveendran}%
  \nametag{\thanks{Akshiya Raveendran and Praveenasarma
  Baskarakurukkal contributed equally to this work.}}
  \Email{raveendiran.21@cse.mrt.ac.lk}\\
  \and
  \Name{Praveenasarma Baskarakurukkal}%
  \nametag{\footnotemark[1]}
  \Email{baskarakurukkal.21@cse.mrt.ac.lk}\\
  \addr University of Moratuwa, Sri Lanka
  \AND
  \Name{Pragatheeswaran Vipulanandan}
  \Email{pxv245@miami.edu}\\
  \and
  \Name{Kamal Premaratne}
  \Email{kamal@miami.edu}\\
  \addr University of Miami, USA
  \AND
  \Name{Uthayasanker Thayasivam}
  \Email{rtuthaya@cse.mrt.ac.lk}\\
  \addr University of Moratuwa, Sri Lanka
}

\begin{document}

\makeatletter
\let \@jmlrpages \@empty
\makeatother

\maketitle

\begin{abstract}
Large Language Models (LLMs) often generate fluent yet incorrect responses, making uncertainty estimation important for reliable deployment. Semantic Entropy (SE) estimates uncertainty by clustering responses based on semantic equivalence, but it relies on hard entailment-based clustering. We propose \textit{Embedding Semantic Entropy}, which replaces entailment-based clustering with K-Medoids in semantic embedding space and uses dynamic cluster selection to account for intra-cluster compactness and inter-cluster separation. We evaluate the method on TriviaQA, SQuAD, NQ-Open, SVAMP, and BioASQ using Falcon-7B-Instruct, LLaMA-2-7B, LLaMA-2-7B-Chat, and Gemma-7B. Results show competitive AUROC and AUARC performance, suggesting that NLI-based entailment clustering can be replaced with embedding-space clustering with limited degradation while improving the inspectability of semantic grouping. This matters because semantic uncertainty scores depend on underlying semantic groups; by making these groups explicit, compact, and separable in embedding space, Emb-SE turns uncertainty estimation from a black-box entailment decision into an inspectable clustering problem.
\end{abstract}


\section{Introduction}

Large Language Models (LLMs) have demonstrated strong performance across a wide range of natural language processing tasks. Despite their fluency and coherence, these models frequently generate answers that are incorrect, fabricated, or unsupported by evidence while expressing high confidence. Such failures, commonly referred to as hallucinations or confabulations, pose serious risks in high-stakes domains where users may over-trust model outputs.

In this work, we focus on short-form question answering (QA), where model outputs are typically concise factual answers such as entities, dates, numerical values, or short statements.

Existing semantic uncertainty methods often rely on pairwise entailment models to group generated answers by meaning. Although effective, this approach introduces an external NLI component, requires many pairwise comparisons, and provides limited visibility into why responses are grouped together. This motivates an alternative formulation in which semantic groups are constructed directly from the geometry of sentence embeddings. Such a formulation can expose cluster compactness, representative medoids, and inter-cluster separation, making the uncertainty score easier to inspect.

\section{Related Works}

A foundational approach in this area is \textit{Semantic Entropy} (SE)~\cite{farquhar2024semantic} which computes uncertainty over semantic meanings rather than surface-level token variations by clustering generated responses into semantically equivalent groups using bidirectional entailment. However, SE relies on hard clustering decisions derived from entailment models and does not capture fine-grained semantic relationships between responses. In particular, it does not explicitly model intra-cluster compactness or inter-cluster separation, which are critical for robust clustering \cite{vipulanandan2026semanticuncertaintyquantificationhallucinations}.

To address the limitations of hard semantic grouping, kernel-based approaches have been proposed~\cite{nikitin2023kle}, which replaces discrete clustering with a kernel-based similarity matrix and computes uncertainty using von Neumann entropy~\cite{bengtsson2017geometry}. This formulation captures pairwise semantic similarities and provides more expressive uncertainty estimates. However, KLE operates on a kernel similarity matrix rather than explicit semantic clusters, it does not provide cluster-level structure such as representative responses, intra-group compactness, or separation between semantic groups. This limits its inspectability: although KLE can indicate that responses are semantically diverse, it does not directly show which semantic groups exist, which response represents each group, or how tightly grouped those responses are.

Quantum-inspired methods extend semantic entropy by modeling token-sequence probability uncertainty with quantum tensor networks and perturbation-based uncertainty quantification~\cite{vipulanandan2026quantum}. While they improve robustness across generation lengths and quantization settings, they mainly calibrate sequence-probability uncertainty and do not explicitly address semantic cluster geometry.

Embedding-space methods such as \textit{Semantic Density}~\cite{qiu2024semanticdensity} estimate uncertainty from the distribution of responses in embedding space, but do not directly incorporate cluster validation or geometric constraints such as intra-cluster compactness and inter-cluster separation.

Beyond single-dimensional approaches, multi-dimensional frameworks combine multiple uncertainty signals, such as semantic similarity between generated responses and knowledge consistency with respect to model outputs or external information~\cite{chen2025multidim}. These signals are computed independently and aggregated to produce a final uncertainty estimate, enabling improved detection of unreliable outputs in complex question answering settings. However, this approach introduces additional computational overhead and system complexity, as it requires multiple scoring components and aggregation mechanisms during inference.

Complementary work has explored hallucination detection using structured representations. Methods such as KEA  ~\cite{haskins2025kea} leverage knowledge graphs and graph kernel techniques to compare generated outputs with structured facts, enabling identification of inconsistencies at the level of explicit knowledge relationships. This provides interpretable signals by highlighting which parts of the generated response align or conflict with known facts. However, such approaches are less directly applicable to uncertainty estimation in short-form answer clustering.

\textbf{Limitations of Existing Works.} Despite these advances, existing methods exhibit several common limitations. First, clustering-based approaches such as SE rely on hard semantic equivalence and fail to capture fine-grained relationships between responses. Second, similarity-based methods such as KLE and Semantic Density do not explicitly enforce cluster geometry or validate clustering quality. These limitations motivate the need for a clustering framework that explicitly uses the geometry of semantic embedding space, including intra-cluster compactness and inter-cluster separation, to improve semantic grouping for uncertainty estimation.

\paragraph{Our Contribution.}
We propose a novel \textit{Embedding Semantic Entropy}, a new embedding-based semantic uncertainty estimator that computes entropy over semantic groups constructed in embedding space rather than through entailment-based clustering.

\begin{itemize}
    \item \textbf{DeBERTa-free semantic clustering:} We replace DeBERTa-based entailment clustering with embedding-based K-Medoids clustering. This shows that a strong external NLI classifier is not always necessary for semantic uncertainty estimation; established clustering techniques in embedding space can sustain, and in several settings improve, uncertainty estimation performance. By relying on standard distance-based clustering rather than pairwise entailment decisions, the proposed method also makes the semantic grouping process more transparent and easier to inspect. We also show that the external DeBERTa-based NLI component used for entailment clustering can be removed without substantial degradation in uncertainty estimation performance.    
   
    \item \textbf{Embedding-based white-box clustering:} Prior NLI-based methods rely on DeBERTa~\cite{he2021deberta}, treating semantic grouping as a black-box decision. In contrast, the proposed method is a white-box approach that provides greater explainability by making the clustering process and cluster structure directly inspectable.

    \item \textbf{Geometry-aware dynamic clustering:} We adapt established cluster-validation criteria to the semantic-response clustering stage. Davies--Bouldin validation~\cite{davies1979cluster} is used to evaluate the trade-off between intra-cluster dispersion and inter-cluster separation, a medoid-based compactness constraint prevents semantically broad clusters, and elbow-based selection~\cite{salvador2004determining} avoids unnecessary over-splitting when additional clusters provide only marginal geometric improvement.

    \item \textbf{Evaluation:} We evaluate the method across QA datasets and open-weight LLMs using AUROC and AUARC.
\end{itemize}
\section{Methodology}

\begin{figure}[t]
\centering
\resizebox{\columnwidth}{!}{%
\begin{tikzpicture}[
    font=\small,
    >=Latex,
    node distance=0.65cm and 0.65cm,
    process/.style={
        draw=blue!80!black,
        rounded corners,
        thick,
        minimum width=3.45cm,
        minimum height=2.45cm,
        align=center,
        fill=blue!3,
        font=\large
    },
    decision/.style={
        diamond,
        draw=teal!70!black,
        thick,
        aspect=1.9,
        align=center,
        inner sep=1.5pt,
        minimum width=3.45cm,
        minimum height=2.45cm,
        fill=teal!5,
        font=\large
    },
    returnbox/.style={
        draw=teal!70!black,
        rounded corners,
        thick,
        minimum width=3.25cm,
        minimum height=2.2cm,
        align=center,
        fill=teal!6,
        font=\large,
    },
    dashedgroup/.style={
        draw=blue!70!black,
        dashed,
        rounded corners,
        thick,
        inner sep=9pt
    },
    arrow/.style={->, thick}
]

% ---------------- Row 1 ----------------
\node[process] (input) {
\textbf{Input:}\\
Question $q$,\\
$n$ generated answers\\
$A=\{a_1,\ldots,a_n\}$
};

\node[process, right=of input] (embed) {
\textbf{Generate}\\
embeddings\\
$\mathbf{e}_i \leftarrow f(a_i')$
};

\node[decision, right=0.75cm of embed] (edge) {
$\max_i \|\hat{\mathbf{e}}_i-\hat{\mathbf{e}}_1\|_2$\\
$< \epsilon_{\mathrm{id}}$?
};

\node[process, right=0.9cm of edge] (onecluster) {
Assign all\\
responses to\\
one cluster
};

\draw[arrow] (input) -- (embed);
\draw[arrow] (embed) -- (edge);
\draw[arrow] (edge) -- node[above]{\textbf{YES}} (onecluster);

% ---------------- Row 2 ----------------
\node[process, below=1.9cm of input] (cluster) {
Cluster embeddings\\
using K-Medoids
};

\node[process, right=of cluster] (db) {
Compute DB index $DB_K$\\
and per-cluster\\
compactness
};

\node[process, right=of db] (select) {
Select optimal $K$ using\\
compactness, DB score,\\
and elbow criterion
};

\node[returnbox, right=of select] (return2) {
Return\\
cluster labels $Y$
};

\node[
    dashedgroup,
    fit=(cluster)(db)(select),
    label={[font=\bfseries\color{blue!70!black}]north:For $K=2$ to $\min(10,n)$}
] (loopbox) {};

\draw[arrow] (edge.south) -- node[right]{\textbf{NO}} ++(0,-0.75) -| (cluster.north);
\draw[arrow] (onecluster.south) -- ++(0,-0.75) -| (return2.north);

\draw[arrow] (cluster) -- (db);
\draw[arrow] (db) -- (select);
\draw[arrow] (select) -- (return2);

\end{tikzpicture}
}
\caption{Flowchart of the proposed embedding-based dynamic clustering algorithm.}
\label{fig:dynamic_clustering_flowchart}
\end{figure}
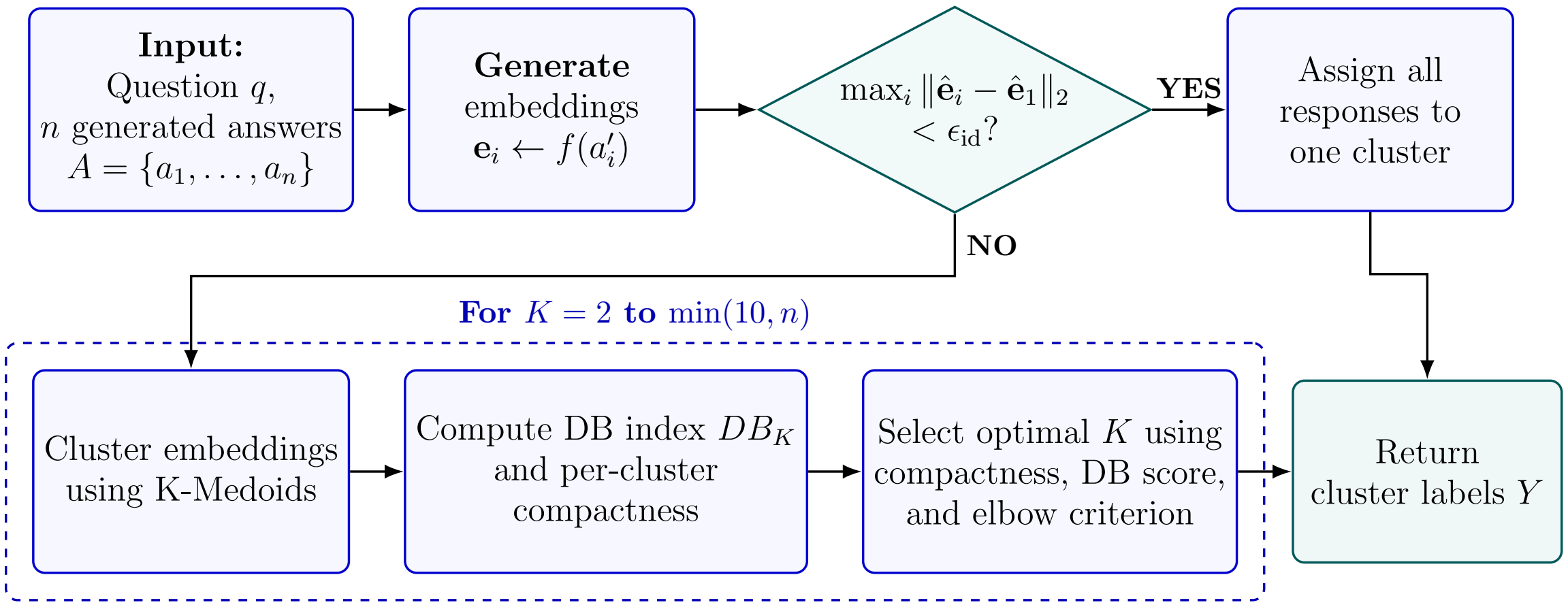

% \subsection{Overview}
Our method improves the clustering stage of Semantic Entropy by replacing entailment-based grouping with
embedding-based clustering. The resulting cluster distribution is then used to compute an entropy-based uncertainty score.
The overall pipeline is summarized in Fig.~\ref{fig:dynamic_clustering_flowchart}.

% \subsection{Input Preparation}
\paragraph{Input Preparation}
Given a question $q$, we sample generated answers $A=\{a_1,\dots,a_n\}$. Each answer is combined with the question before embedding to preserve contextual meaning, i.e., $a_i'=\text{``Context: }q\text{ \textbackslash n Answer: }a_i$. This context-aware formatting is important for short-form QA. A bare answer such as ``London'' may be ambiguous without the question, whereas the formatted input represents the answer in relation to the question being answered. We then apply basic preprocessing, including lowercasing, article removal, Unicode normalization, and noise cleaning, to reduce lexical variability.

% \subsection{Embedding Representation}
\paragraph{Embedding Representation}
Each processed answer is encoded into a dense semantic vector $\mathbf{e}_i = f(a_i')$, where $f(\cdot)$ denotes a sentence embedding
encoder. In our implementation, we use supervised SimCSE with RoBERTa-large,
\url{princeton-nlp/sup-simcse-roberta-large}~\cite{gao2021simcse,princeton2021supsimcse}. We select supervised SimCSE because the clustering stage requires an embedding space in which semantically similar answers are placed close together while semantically different answers are separated. Compared with generic sentence encoders, supervised SimCSE is explicitly trained using contrastive learning to improve sentence-level semantic similarity, making it suitable for grouping short generated answers. It also provides fixed-dimensional dense embeddings that can be efficiently clustered using distance-based methods such as K-Medoids. This is preferable for our setting because the clustering procedure depends directly on the geometry of the embedding space rather than on pairwise entailment labels. We then L2-normalize each embedding as $\hat{\mathbf{e}}_i = \mathbf{e}_i / \|\mathbf{e}_i\|$.

\subsection{Degeneracy Handling}

Before the main clustering loop, we handle two degenerate cases. If $n=1$, the only response is assigned to cluster 0. If all normalized embeddings are near-identical, all responses are also assigned to a single cluster. We detect this using
\begin{equation}
\max_i \|\hat{\mathbf{e}}_i-\hat{\mathbf{e}}_1\|_2 < \epsilon_{\mathrm{id}}.
\end{equation}
 We use $\epsilon_{\mathrm{id}}=10^{-8}$ only as a numerical tolerance for detecting duplicate or near-duplicate embeddings. Since embeddings are L2-normalized, Euclidean distance is directly related to cosine similarity:
\begin{equation}
\|\hat{\mathbf{e}}_i-\hat{\mathbf{e}}_j\|_2^2
=
2 - 2\cos(\hat{\mathbf{e}}_i,\hat{\mathbf{e}}_j).
\end{equation}
Therefore, when the distance is below $10^{-8}$, the corresponding cosine similarity is essentially 1 up to floating-point precision. In this edge case, separating the responses into multiple clusters would be unstable and unnecessary. This step avoids unnecessary clustering when all generated answers have the same meaning. In such cases, all responses are assigned to one cluster, indicating zero semantic disagreement.

\subsection{Clustering with K-Medoids}

For non-degenerate cases, we cluster the normalized embeddings using K-Medoids. For each candidate number of clusters
\begin{equation}
K \in \{2,\dots,\min(10,n)\},
\end{equation}
Since typical experiments use 10 sampled responses per question, $K=10$ is the theoretical maximum, corresponding to the case where each response forms its own cluster. However, cluster-validity metrics such as the Davies--Bouldin index are not meaningful when every point is assigned to a separate singleton cluster. Therefore, candidate clusterings are evaluated only when they produce at least two and fewer than $n$ non-empty clusters. This allows the method to explore meaningful semantic partitions while avoiding degenerate all-singleton solutions.

We use K-Medoids where the cluster representative is an actual generated answer. This makes each cluster more interpretable: the medoid can be inspected as the representative response for that semantic group. K-Means centroids, in contrast, are arithmetic averages that may not correspond to any real response and may fall inside the unit hypersphere after averaging normalized vectors.

In our implementation, K-Medoids uses PAM-style optimization~\cite{schubert2019fast}, k-medoids++ initialization, Euclidean distance, and a fixed random seed.  This choice is important because the clusters produced at this stage directly determine the probability masses used in the final entropy calculation. PAM-style optimization improves the quality of medoid selection by reducing within-cluster distance, which helps form compact semantic groups. The k-medoids++ initialization further improves stability by starting from well-separated candidate medoids, reducing sensitivity to poor initialization. Together, these choices make the semantic clusters more reliable and reproducible, which is essential because over-merging distinct meanings would underestimate uncertainty, while over-splitting paraphrases would overestimate uncertainty.

\paragraph{Dynamic Selection of Number of Clusters}

Selecting an appropriate number of clusters is critical for capturing the semantic structure of generated responses. Instead of fixing the number of clusters \textit{a priori}, we evaluate multiple candidate values to identify the clustering configuration that best separates distinct answer meanings while keeping semantically equivalent responses within compact groups.

\paragraph{Rationale for Geometry-Aware Validation.}
The goal of the clustering stage is not to introduce a new clustering validity index, but to use established geometric criteria to construct reliable semantic groups for entropy estimation. Since the responses are represented as L2-normalized embeddings, Euclidean distances reflect cosine-based semantic similarity. In our setting, each generated response is represented as an L2-normalized embedding vector. Therefore, the structure of the embedding space can be analyzed through pairwise distances between response embeddings, distances from responses to their cluster medoids, and distances between cluster medoids. These distances are important because they determine whether responses assigned to the same cluster are semantically close and whether different clusters represent meaningfully different answer groups. If distinct meanings are merged into one loose cluster, the final entropy score may underestimate uncertainty. Conversely, if paraphrases are split into separate clusters, the entropy score may overestimate uncertainty. We therefore use distance-based validation to favor clusters that are internally compact and externally separated before computing the final entropy score.

\paragraph{Davies--Bouldin Index.} For each candidate $K$, we compute the Davies--Bouldin (DB) index~\cite{davies1979cluster}:
\begin{equation}
DB = \frac{1}{K} \sum_{i=1}^{K} \max_{j \neq i} \frac{S_i + S_j}{M_{ij}},
\end{equation}
where $S_i$ denotes the intra-cluster dispersion of cluster $i$, and $M_{ij}$ represents the distance between clusters $i$ and $j$. The DB index jointly captures intra-cluster compactness and inter-cluster separation, with lower values indicating better clustering quality.

Davies--Bouldin validation is suitable because it jointly measures intra-cluster dispersion and inter-cluster separation without requiring ground-truth cluster labels~\cite{davies1979cluster}. This is important in our setting because the true semantic clusters of sampled LLM responses are not known during inference. However, the Davies--Bouldin score alone can hide poor individual clusters because it summarizes clustering quality globally~\cite{arbelaitz2013extensive}. We therefore add an explicit medoid-based compactness constraint to reject any candidate clustering that contains a loose semantic group.

\paragraph{Explicit Compactness Constraint.} To address the limitation of relying only on a global DB score, we introduce an explicit compactness constraint that evaluates clustering quality at the level of individual clusters. For each cluster $C$, we compute the mean distance between its points and the corresponding medoid:
\begin{equation}
D_C = \frac{1}{|C|} \sum_{x \in C} d(x, m_C),
\end{equation}
where $m_C$ is the medoid of cluster $C$, and $d(x,m_C)$ denotes the Euclidean distance between response embedding $x$ and medoid $m_C$ in the L2-normalized embedding space. A clustering configuration is rejected if any cluster violates the compactness threshold, i.e., $\max_C D_C > \tau$. This prevents the merging of semantically dissimilar responses into overly broad clusters, which would otherwise reduce the final entropy score by assigning distinct meanings to the same semantic group.

\paragraph{Elbow-Based Selection~\cite{salvador2004determining}} Among the remaining valid candidates, we analyze the rate of improvement in DB scores to avoid selecting unnecessarily large values of $K$. For $K=1$, there is only one cluster, so the DB score and elbow comparison are not meaningful because no inter-cluster separation exists. For $K=2$, the DB score can be computed, but the elbow ratio cannot be formed because it requires three consecutive DB scores. For $K \geq 3$, let:
\begin{equation}
\Delta_1 = DB_{K-2} - DB_{K-1}, \quad
\Delta_2 = DB_{K-1} - DB_K.
\end{equation}
We define the improvement ratio as:
\begin{equation}
r = \frac{|\Delta_2|}{|\Delta_1| + \epsilon},
\end{equation}
where $\epsilon$ is a small constant for numerical stability. A sharp decrease in this ratio ($r$) indicates diminishing returns from increasing $K$. When $r < \gamma$, where $\gamma$ is the elbow sensitivity threshold, we identify an elbow point and select the previous number of clusters, i.e., $K^* = K - 1$.
 The optimal number of clusters is therefore determined using a multi-stage procedure. Candidate values of $K$ that violate the compactness constraint are first discarded. The elbow criterion is then applied to the remaining candidates to detect diminishing improvements in DB scores. If no clear elbow point is observed, the value of $K$ that minimizes the DB index among the valid candidates is selected as a fallback. This fallback is justified because, after removing candidates with loose clusters, the remaining candidates already satisfy the compactness requirement; among them, the minimum DB score corresponds to the best available trade-off between intra-cluster compactness and inter-cluster separation. This procedure favors clusterings that are compact, separated, and not unnecessarily over-split, making the final semantic groups more reliable for entropy estimation.

\paragraph{Why Geometry-Aware Clustering is Needed.}
A direct replacement of NLI clustering with embedding clustering is not sufficient for semantic uncertainty estimation, because the number and quality of semantic groups directly determine the final entropy score. If embedding-based clustering merges distinct meanings into one broad cluster, the entropy score is underestimated; if it over-splits paraphrases into many small clusters, the entropy score is overestimated. Therefore, the clustering stage must not only group responses by embedding similarity, but also control the geometry of the resulting semantic groups. Emb-SE addresses this by combining medoid-based clustering, explicit compactness checking, Davies--Bouldin validation, and elbow-based model selection. These components serve complementary roles: K-Medoids provides representative generated answers as cluster centers, the compactness constraint prevents semantically loose clusters, the Davies--Bouldin index balances intra-cluster dispersion against inter-cluster separation, and the elbow criterion avoids unnecessary over-splitting. Together, this makes Emb-SE more than a straightforward embedding-based replacement of NLI clustering; it provides a geometry-aware mechanism for constructing semantic groups whose structure directly affects the uncertainty estimate.

\subsection{Hyperparameter Configuration}

We tune the clustering hyperparameters using a 5-fold cross-validation grid search for each model--dataset setting. The search is performed over the compactness threshold $\tau \in [0.10,0.50]$ and the elbow sensitivity threshold $\gamma \in [0.01,0.50]$. In the implementation, $\tau$ is searched from 0.10 to 0.50 with step size 0.05, while $\gamma$ is searched from 0.01 to 0.50 with step size 0.01. For each candidate configuration, the validation responses are clustered, the Embedding Semantic Entropy score is computed, and the configuration is evaluated using AUROC. The configuration with the highest mean cross-validation AUROC is selected for the corresponding model--dataset pair.

Appendix~\ref{app:hyperparameter_selection}, Table~\ref{tab:hyperparameter_summary} summarizes the selected hyperparameter values. The compactness threshold $\tau$ is summarized by model, since it is mainly affected by model-specific response dispersion. The elbow sensitivity threshold $\gamma$ is summarized by dataset, since it reflects how clearly responses separate into semantic clusters for a given benchmark. Overall, Falcon-7B-Instruct tends to select stricter compactness values, while LLaMA-2-7B-Chat selects more relaxed compactness values. The selected $\gamma$ values vary more widely across datasets, indicating that the elbow behavior of the Davies--Bouldin curve depends on the structure of the answer space.

\subsection{Uncertainty Estimation}

After the final clustering step, we compute an entropy-based uncertainty score over the embedding-derived semantic clusters. The key idea is to first compute the probability of each generated sequence, then aggregate these probabilities within each semantic cluster, and finally compute entropy over the resulting cluster-level probability distribution.

Let $q$ denote the input question and let
$A=\{a_1,\dots,a_n\}$ be the sampled responses. We denote the token sequence corresponding to response $a_i$ as
\begin{equation}
S_i = (t_{i,1}, t_{i,2}, \dots, t_{i,N_i}),
\end{equation}
where $i$ indexes the generated response, $N_i$ is the number of generated tokens in response $a_i$. The model assigns a conditional probability to each token given the question and the previously generated tokens:
\begin{equation}
p_{i,r}
=
p(t_{i,r} \mid q, t_{i,<r}).
\end{equation}
where $r$ indexes the token position within that response. Then, we compute the sequence log-likelihood as
\begin{equation}
\ell_i
=
\sum_{r=1}^{N_i} \log p_{i,r}.
\end{equation}
In practice, we use log-likelihoods for numerical stability, since multiplying many token probabilities can produce very small values.

The raw sequence probabilities are then normalized over the sampled responses:
\begin{equation}
P_{S_i}
=
\frac{\exp(\ell_i)}
{\sum_{j=1}^{n} \exp(\ell_j)}.
\end{equation}
Here, $P_{S_i}$ is the normalized probability mass assigned to sampled sequence $S_i$, and $\sum_{i=1}^{n} P_{S_i}=1$.

Let $y_i \in \{1,\dots,K^*\}$ denote the final cluster assignment of response $a_i$, where $K^*$ is the selected number of clusters. Each cluster is defined as a set of response indices:
\begin{equation}
C_k = \{i : y_i = k\}.
\end{equation}
The probability mass of cluster $C_k$ is obtained by summing the normalized sequence probabilities of all responses assigned to that cluster:
\begin{equation}
P_{C_k}
=
\sum_{i \in C_k} P_{S_i}.
\end{equation}

This aggregation prevents surface-level paraphrases from being treated as independent semantic outcomes.

The cluster probabilities are also normalized:
\begin{equation}
\sum_{k=1}^{K^*} P_{C_k}(q)
=
\sum_{k=1}^{K^*}
\sum_{i \in C_k} P_{S_i}(q)
=
\sum_{i=1}^{n} P_{S_i}(q)
=
1.
\end{equation}
Therefore, the cluster probabilities form a valid probability distribution over the semantic clusters.

The Embedding Semantic Entropy score is then computed over the cluster probability distribution:
\begin{equation}
H_{\mathrm{EmbSE}}(q)
=
-\sum_{k=1}^{K^*}
P_{C_k}(q)\log P_{C_k}(q).
\end{equation}

Low entropy indicates that most probability mass is concentrated in one embedding-derived semantic cluster, suggesting high confidence in a single semantic answer. High entropy indicates that probability mass is spread across multiple semantic clusters, reflecting greater semantic disagreement among the sampled responses.

Thus, Emb-SE estimates semantic uncertainty over groups rather than individual answer strings by using geometry-aware clustering in embedding space. The proposed method constructs these semantic groups using embedding-space distances, medoid-based compactness, and inter-cluster separation before computing the final entropy-based uncertainty score.

\section{Experiments}

We evaluate on short-form QA by generating multiple responses per question and estimating semantic uncertainty to identify unreliable outputs. 

\paragraph{Datasets and Models.}
We evaluate on five short-form QA benchmarks: TriviaQA~\cite{joshi2017triviaqa}, SQuAD~\cite{rajpurkar2016squad}, NQ-Open~\cite{lee2019latent}, SVAMP~\cite{patel2021nlp}, and BioASQ~\cite{tsatsaronis2015bioasq}. These datasets cover open-domain factual question answering, reading comprehension, arithmetic word problems, and biomedical question answering, allowing us to test whether the proposed clustering method is robust across different answer distributions.

Experiments are conducted using four open-weight LLMs: Falcon-7B-Instruct~\cite{falcon2023}, LLaMA-2-7B, LLaMA-2-7B-Chat~\cite{touvron2023llama2}, and Gemma-7B~\cite{team2024gemma}. For each input question, $n = 10$ responses are sampled at temperature $T = 1.0$ using top-$K$ sampling ($K = 50$) . An additional low-temperature sample ($T = 0.1$) is used to assess model accuracy against ground-truth answers. The exact answer-generation prompt format is provided in Appendix~\ref{app:generation_prompt}.

\paragraph{Baselines.} We compare against a range of uncertainty estimation methods spanning semantic, probabilistic, and embedding-based approaches. These include Semantic Entropy (SE), Discrete Semantic Entropy (DSE), and Embedding Regression (ER)~\cite{farquhar2024semantic}, Kernel Language Entropy (KLE)~\cite{nikitin2023kle}, Predictive Entropy (PE)~\cite{malinin2020uncertainty}, and P(True)~\cite{kadavath2022language}. We additionally evaluate our proposed method, Embedding Semantic Entropy.

\paragraph{Evaluation Metrics.} We evaluate performance using Area Under the Receiver Operating Characteristic curve (AUROC) and Area Under the Accuracy-Rejection Curve (AUARC)~\cite{nadeem2009arc}. These metrics are commonly used in state-of-the-art uncertainty estimation and hallucination detection evaluations~\cite{farquhar2024semantic,nikitin2023kle}. AUROC evaluates the ability to distinguish correct and incorrect predictions, whereas AUARC evaluates whether uncertainty estimates are useful for rejecting low-confidence predictions.

\subsection{Results}

Figs.~\ref{fig:auroc_pairwise_winrate} and~\ref{fig:auarc_pairwise_winrate} summarize the pairwise win-rate comparison across all 20 model--dataset settings, corresponding to four LLMs evaluated on five QA datasets. The complete numerical AUROC and AUARC results are reported in Appendix~\ref{app:full_results}, with model-wise results shown in Tables~\ref{tab:falcon7b-instruct-auroc-auarc}--\ref{tab:gemma7b-auroc-auarc}. Overall, Embedding Semantic Entropy performs competitively against strong semantic, probabilistic, and embedding-based uncertainty baselines, and is frequently the tied-best method.

Across the 20 model--dataset settings, Emb-SE achieves the best or tied-best AUROC in 15 settings and the best or tied-best AUARC in 13 settings. This indicates that the proposed embedding-based clustering method is not only competitive with entailment-based Semantic Entropy, but also consistently strong across diverse datasets and model families. The strongest performance is observed for Gemma-7B, where Emb-SE achieves the best or tied-best AUARC on all five datasets and the best AUROC on four of the five datasets. Emb-SE also performs strongly for Falcon-7B-Instruct and LLaMA-2-7B, especially on TriviaQA, NQ, and BioASQ.

Fig.~\ref{fig:auroc_pairwise_winrate} shows the pairwise AUROC win-rate heatmap. Each cell reports the fraction of settings in which the row method outperforms the column method, with ties counted as 0.5. Emb-SE obtains the strongest overall pairwise AUROC profile. It outperforms or ties SE with a pairwise win rate of 0.80, DSE with 0.72, KLE($K_{\mathrm{HEAT}}$) with 0.90, PE with 0.90, and both $P(\mathrm{True})$ and ER with 1.00.

The appendix tables provide a more detailed model-wise view. For Falcon-7B-Instruct, Emb-SE obtains the highest AUROC on TriviaQA, SQuAD, NQ, and BioASQ, while PE is strongest on SVAMP. For LLaMA-2-7B, Emb-SE is best on TriviaQA, NQ, and BioASQ, while DSE is strongest on SQuAD and SE is strongest on SVAMP. For LLaMA-2-7B-Chat, Emb-SE is best or tied-best on TriviaQA, SQuAD, NQ, and BioASQ, but weaker on SVAMP. For Gemma-7B, Emb-SE achieves the highest AUROC on TriviaQA, SQuAD, SVAMP, and BioASQ, while DSE is strongest on NQ.

These results suggest that embedding-space clustering provides discriminative semantic uncertainty scores comparable to, and often stronger than, entailment-based semantic grouping. However, performance still varies by dataset and model, showing that no single uncertainty estimator dominates every setting.

Fig.~\ref{fig:auarc_pairwise_winrate} shows the pairwise AUARC win-rate heatmap. Emb-SE again has the strongest overall profile, with a pairwise AUARC win rate of 0.70 against SE, 0.95 against DSE, 0.90 against KLE($K_{\mathrm{HEAT}}$), 0.80 against PE, 1.00 against $P(\mathrm{True})$, and 0.95 against ER. This indicates that Emb-SE is especially effective in rejection-based evaluation, where the uncertainty score is used to rank unreliable responses for abstention.

The appendix tables show that Emb-SE achieves particularly strong AUARC performance for Gemma-7B, where it is best or tied-best on all five datasets. For LLaMA-2-7B, Emb-SE is best on TriviaQA, NQ, and BioASQ, while PE is strongest on SQuAD and SE is strongest on SVAMP. For LLaMA-2-7B-Chat, Emb-SE is best on TriviaQA, SQuAD, and BioASQ, while SE is slightly stronger on NQ and KLE($K_{\mathrm{HEAT}}$) is strongest on SVAMP. For Falcon-7B-Instruct, Emb-SE is best or tied-best on TriviaQA and NQ, while PE and ER remain stronger on some other datasets.

These results indicate that Emb-SE is useful for selective prediction: it can effectively identify outputs that should be rejected when the model is uncertain.

\paragraph{Comparison with Entailment-Based Approaches.}
Compared with SE and DSE, Emb-SE removes the external DeBERTa-based pairwise entailment clustering step while remaining competitive or stronger in many settings. The pairwise heatmaps show that Emb-SE has a higher AUROC win rate than both SE and DSE, and a higher AUARC win rate than both SE and DSE. This supports the central motivation of the proposed method: semantic uncertainty estimation can be performed using geometry-aware clustering in embedding space, without requiring pairwise NLI inference during uncertainty estimation.

A key advantage of Emb-SE is that the uncertainty score is computed over clusters formed in a continuous embedding space. Unlike entailment-based methods, where grouping depends on discrete NLI labels, the proposed method exposes the geometry behind each semantic group. Cluster medoids, response-to-medoid distances, inter-cluster distances, and compactness values can be inspected directly. This makes the resulting confidence or uncertainty score more interpretable: high uncertainty can be traced to probability mass being spread across well-separated semantic clusters, while low uncertainty corresponds to probability mass concentrated in a compact cluster. Thus, Emb-SE not only maintains competitive uncertainty estimation performance, but also provides a more explainable clustering structure than black-box entailment-based grouping.

At the same time, the appendix tables show that entailment-based or probabilistic baselines remain stronger in selected cases, especially on some SQuAD and SVAMP settings. Therefore, Emb-SE should not be interpreted as universally superior across all benchmarks. Rather, it provides a practical and more inspectable alternative to hard entailment-based clustering, while maintaining strong empirical performance.

Overall, the results show that Emb-SE is a competitive semantic uncertainty estimator across five QA datasets and four open-weight LLMs. The pairwise win-rate heatmaps show that Emb-SE has the strongest overall comparison profile for both AUROC and AUARC, while the appendix tables confirm that this performance is consistent across several models and datasets. These findings support the use of geometry-aware embedding clustering as a transparent alternative to DeBERTa-based semantic grouping.

\vspace{0.3em}
% ============================================================
% Pairwise win-rate heatmap helper macros
% ============================================================

\newcommand{\HeatmapCellFont}{\small\bfseries\sffamily}
\newcommand{\HeatmapLabelFont}{\small\sffamily}
\newcommand{\HeatmapBarFont}{\small\sffamily}

% ColorBrewer RdBu palette: red -> white -> blue
\definecolor{rdBu0}{HTML}{67001F}
\definecolor{rdBu1}{HTML}{B2182B}
\definecolor{rdBu2}{HTML}{D6604D}
\definecolor{rdBu3}{HTML}{F4A582}
\definecolor{rdBu4}{HTML}{FDDBC7}
\definecolor{rdBu5}{HTML}{F7F7F7}
\definecolor{rdBu6}{HTML}{D1E5F0}
\definecolor{rdBu7}{HTML}{92C5DE}
\definecolor{rdBu8}{HTML}{4393C3}
\definecolor{rdBu9}{HTML}{2166AC}
\definecolor{rdBu10}{HTML}{053061}

% Diagonal cross-hatch color
\definecolor{diagblue}{HTML}{8BD3E6}

\newcommand{\CellColor}[1]{%
  \ifdim#1pt<0.15pt
    \xdef\cellfill{rdBu0}%
  \else\ifdim#1pt<0.25pt
    \xdef\cellfill{rdBu1}%
  \else\ifdim#1pt<0.35pt
    \xdef\cellfill{rdBu2}%
  \else\ifdim#1pt<0.45pt
    \xdef\cellfill{rdBu3}%
  \else\ifdim#1pt<0.50pt
    \xdef\cellfill{rdBu4}%
  \else\ifdim#1pt<0.55pt
    \xdef\cellfill{rdBu5}%
  \else\ifdim#1pt<0.65pt
    \xdef\cellfill{rdBu6}%
  \else\ifdim#1pt<0.75pt
    \xdef\cellfill{rdBu7}%
  \else\ifdim#1pt<0.85pt
    \xdef\cellfill{rdBu8}%
  \else\ifdim#1pt<0.95pt
    \xdef\cellfill{rdBu9}%
  \else
    \xdef\cellfill{rdBu10}%
  \fi\fi\fi\fi\fi\fi\fi\fi\fi\fi
}

\newcommand{\WinCell}[3]{%
  \CellColor{#3}%
  \draw[draw=white, line width=0.5pt, fill=\cellfill] (#1,#2) rectangle ++(1,1);
  \node[text=white, font=\HeatmapCellFont] at (#1+0.5,#2+0.5) {#3};
}

\newcommand{\DiagCell}[2]{%
  \draw[draw=white, line width=0.5pt, fill=diagblue] (#1,#2) rectangle ++(1,1);
  \path[pattern=north east lines, pattern color=white] (#1,#2) rectangle ++(1,1);
  \path[pattern=north west lines, pattern color=white] (#1,#2) rectangle ++(1,1);
}

\newcommand{\HeatmapColorbar}{%
  \begin{scope}[shift={(7.55,0)}]
    \foreach \k in {0,...,100}{
      \pgfmathsetmacro{\v}{\k/100}
      \pgfmathsetmacro{\yone}{7*\k/100}
      \pgfmathsetmacro{\ytwo}{7*(\k+1)/100}
      \CellColor{\v}
      \draw[draw=\cellfill, fill=\cellfill] (0,\yone) rectangle (0.25,\ytwo);
    }
    \draw[black, line width=0.3pt] (0,0) rectangle (0.25,7);
    \foreach \v/\lab in {0.2/{0.2},0.4/{0.4},0.6/{0.6},0.8/{0.8}}{
      \node[anchor=west, font=\scriptsize] at (0.35,7*\v) {\lab};
    }
    \node[rotate=90, font=\scriptsize] at (0.95,3.5) {Win rate};
  \end{scope}
}

\newcommand{\HeatmapLabels}{%
  \foreach \i/\lab in {
    0/{Emb-SE},
    1/{SE},
    2/{DSE},
    3/{KLE($K_{\mathrm{HEAT}}$)},
    4/{PE},
    5/{$P(\mathrm{True})$},
    6/{ER}
  }{
    \node[rotate=45, anchor=west, font=\scriptsize] at (\i+0.18,7.15) {\lab};
  }

  \foreach \i/\lab in {
    0/{Emb-SE},
    1/{SE},
    2/{DSE},
    3/{KLE($K_{\mathrm{HEAT}}$)},
    4/{PE},
    5/{$P(\mathrm{True})$},
    6/{ER}
  }{
    \pgfmathtruncatemacro{\yy}{6-\i}
    \node[anchor=east, font=\scriptsize] at (-0.15,\yy+0.5) {\lab};
  }
}

\renewcommand{\HeatmapColorbar}{%
  \begin{scope}[shift={(7.55,0)}]
    \foreach \k in {0,...,100}{
      \pgfmathsetmacro{\v}{\k/100}
      \pgfmathsetmacro{\yone}{7*\k/100}
      \pgfmathsetmacro{\ytwo}{7*(\k+1)/100}
      \CellColor{\v}
      \draw[draw=\cellfill, fill=\cellfill] (0,\yone) rectangle (0.25,\ytwo);
    }
    \draw[black, line width=0.3pt] (0,0) rectangle (0.25,7);
    \foreach \v/\lab in {0.2/{0.2},0.4/{0.4},0.6/{0.6},0.8/{0.8}}{
      \node[anchor=west, font=\scriptsize] at (0.35,7*\v) {\lab};
    }
    \node[rotate=90, font=\scriptsize] at (0.95,3.5) {Win rate};
  \end{scope}
}

% ============================================================
% AUROC pairwise win-rate heatmap
% ============================================================

\begin{figure}[t]
\centering
\resizebox{\columnwidth}{!}{%
\begin{tikzpicture}

% Row 1: Emb-SE
\DiagCell{0}{6}
\WinCell{1}{6}{0.80}
\WinCell{2}{6}{0.72}
\WinCell{3}{6}{0.90}
\WinCell{4}{6}{0.90}
\WinCell{5}{6}{1.00}
\WinCell{6}{6}{1.00}

% Row 2: SE
\WinCell{0}{5}{0.20}
\DiagCell{1}{5}
\WinCell{2}{5}{0.55}
\WinCell{3}{5}{0.75}
\WinCell{4}{5}{0.92}
\WinCell{5}{5}{1.00}
\WinCell{6}{5}{1.00}

% Row 3: DSE
\WinCell{0}{4}{0.28}
\WinCell{1}{4}{0.45}
\DiagCell{2}{4}
\WinCell{3}{4}{0.85}
\WinCell{4}{4}{0.90}
\WinCell{5}{4}{1.00}
\WinCell{6}{4}{0.95}

% Row 4: KLE
\WinCell{0}{3}{0.10}
\WinCell{1}{3}{0.25}
\WinCell{2}{3}{0.15}
\DiagCell{3}{3}
\WinCell{4}{3}{0.68}
\WinCell{5}{3}{0.95}
\WinCell{6}{3}{0.90}

% Row 5: PE
\WinCell{0}{2}{0.10}
\WinCell{1}{2}{0.08}
\WinCell{2}{2}{0.10}
\WinCell{3}{2}{0.32}
\DiagCell{4}{2}
\WinCell{5}{2}{0.95}
\WinCell{6}{2}{0.90}

% Row 6: P(True)
\WinCell{0}{1}{0.00}
\WinCell{1}{1}{0.00}
\WinCell{2}{1}{0.00}
\WinCell{3}{1}{0.05}
\WinCell{4}{1}{0.05}
\DiagCell{5}{1}
\WinCell{6}{1}{0.55}

% Row 7: ER
\WinCell{0}{0}{0.00}
\WinCell{1}{0}{0.00}
\WinCell{2}{0}{0.05}
\WinCell{3}{0}{0.10}
\WinCell{4}{0}{0.10}
\WinCell{5}{0}{0.45}
\DiagCell{6}{0}

\HeatmapLabels
\HeatmapColorbar

\end{tikzpicture}
}
\caption{Pairwise AUROC win-rate heatmap across 20 model--dataset settings. Each cell shows the fraction of settings in which the row method outperforms the column method.}
\label{fig:auroc_pairwise_winrate}
\end{figure}
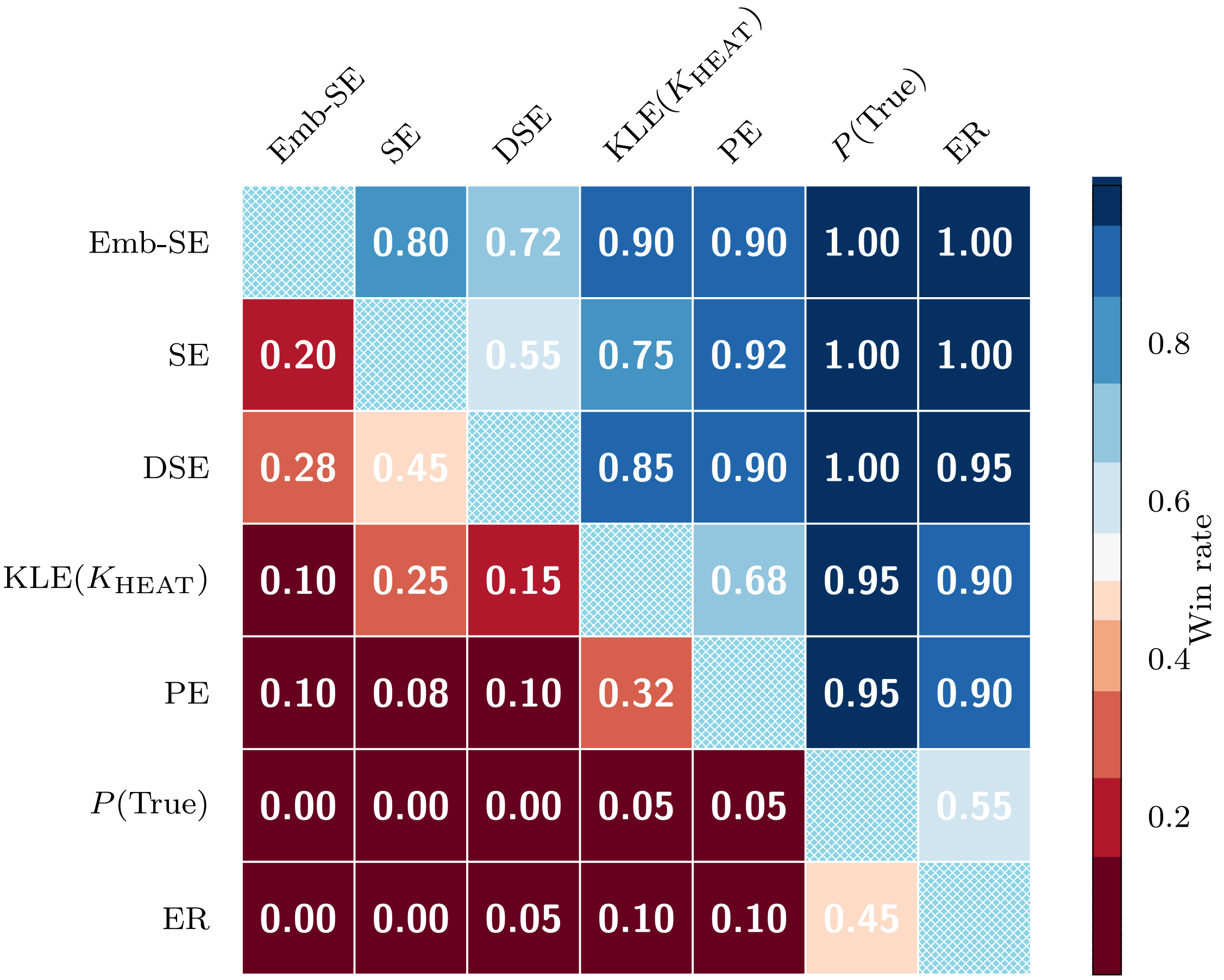

% ============================================================
% AUARC pairwise win-rate heatmap
% ============================================================

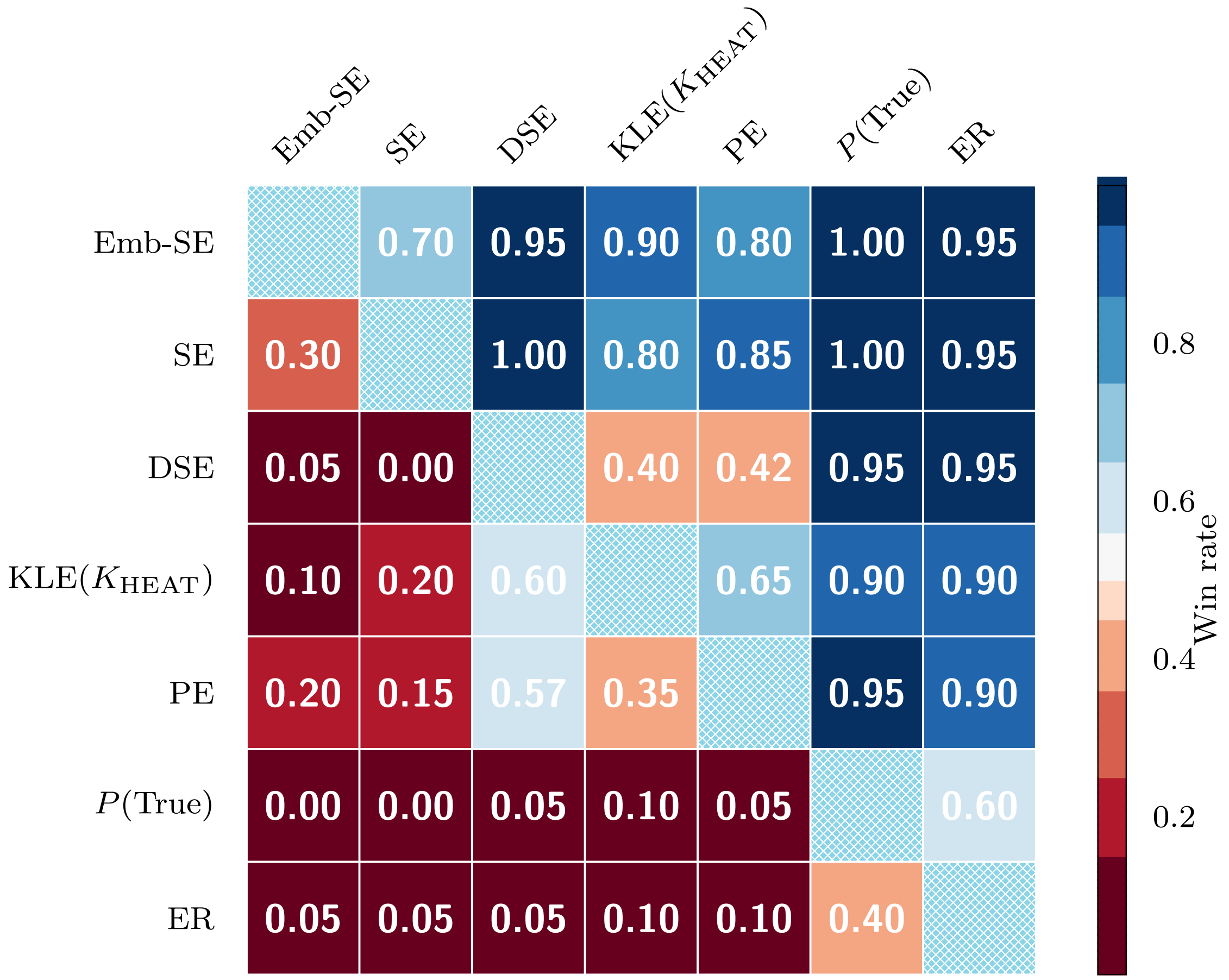
\begin{figure}[t]
\centering
\resizebox{\columnwidth}{!}{%
\begin{tikzpicture}

% Row 1: Emb-SE
\DiagCell{0}{6}
\WinCell{1}{6}{0.70}
\WinCell{2}{6}{0.95}
\WinCell{3}{6}{0.90}
\WinCell{4}{6}{0.80}
\WinCell{5}{6}{1.00}
\WinCell{6}{6}{0.95}

% Row 2: SE
\WinCell{0}{5}{0.30}
\DiagCell{1}{5}
\WinCell{2}{5}{1.00}
\WinCell{3}{5}{0.80}
\WinCell{4}{5}{0.85}
\WinCell{5}{5}{1.00}
\WinCell{6}{5}{0.95}

% Row 3: DSE
\WinCell{0}{4}{0.05}
\WinCell{1}{4}{0.00}
\DiagCell{2}{4}
\WinCell{3}{4}{0.40}
\WinCell{4}{4}{0.42}
\WinCell{5}{4}{0.95}
\WinCell{6}{4}{0.95}

% Row 4: KLE
\WinCell{0}{3}{0.10}
\WinCell{1}{3}{0.20}
\WinCell{2}{3}{0.60}
\DiagCell{3}{3}
\WinCell{4}{3}{0.65}
\WinCell{5}{3}{0.90}
\WinCell{6}{3}{0.90}

% Row 5: PE
\WinCell{0}{2}{0.20}
\WinCell{1}{2}{0.15}
\WinCell{2}{2}{0.57}
\WinCell{3}{2}{0.35}
\DiagCell{4}{2}
\WinCell{5}{2}{0.95}
\WinCell{6}{2}{0.90}

% Row 6: P(True)
\WinCell{0}{1}{0.00}
\WinCell{1}{1}{0.00}
\WinCell{2}{1}{0.05}
\WinCell{3}{1}{0.10}
\WinCell{4}{1}{0.05}
\DiagCell{5}{1}
\WinCell{6}{1}{0.60}

% Row 7: ER
\WinCell{0}{0}{0.05}
\WinCell{1}{0}{0.05}
\WinCell{2}{0}{0.05}
\WinCell{3}{0}{0.10}
\WinCell{4}{0}{0.10}
\WinCell{5}{0}{0.40}
\DiagCell{6}{0}

\HeatmapLabels
\HeatmapColorbar

\end{tikzpicture}
}
\caption{Pairwise AUARC win-rate heatmap across 20 model--dataset settings. Each cell shows the fraction of settings in which the row method outperforms the column method; ties are counted as 0.5.}
\label{fig:auarc_pairwise_winrate}
\end{figure}

\newcommand{\SignedCell}[4]{%
  \CellColor{#3}%
  \draw[draw=white, line width=0.5pt, fill=\cellfill] (#1,#2) rectangle ++(1,1);
  \def\celltext{black}%
  \ifdim#3pt<0.35pt
    \def\celltext{white}%
  \fi
  \ifdim#3pt>0.72pt
    \def\celltext{white}%
  \fi
  \node[text=\celltext, font=\small] at (#1+0.5,#2+0.5) {#4};
}

\begin{figure}[t]
\centering
\resizebox{\columnwidth}{!}{%
\begin{tikzpicture}

% Row 1: Emb-SE
\DiagCell{0}{6}
\SignedCell{1}{6}{0.505}{0.005}
\SignedCell{2}{6}{0.506}{0.006}
\SignedCell{3}{6}{0.525}{0.025}
\SignedCell{4}{6}{0.542}{0.042}
\SignedCell{5}{6}{0.647}{0.147}
\SignedCell{6}{6}{0.627}{0.127}

% Row 2: SE
\SignedCell{0}{5}{0.495}{-0.005}
\DiagCell{1}{5}
\SignedCell{2}{5}{0.500}{0.000}
\SignedCell{3}{5}{0.520}{0.020}
\SignedCell{4}{5}{0.537}{0.037}
\SignedCell{5}{5}{0.642}{0.142}
\SignedCell{6}{5}{0.622}{0.122}

% Row 3: DSE
\SignedCell{0}{4}{0.494}{-0.006}
\SignedCell{1}{4}{0.500}{0.000}
\DiagCell{2}{4}
\SignedCell{3}{4}{0.520}{0.020}
\SignedCell{4}{4}{0.537}{0.037}
\SignedCell{5}{4}{0.641}{0.141}
\SignedCell{6}{4}{0.622}{0.122}

% Row 4: KLE
\SignedCell{0}{3}{0.475}{-0.025}
\SignedCell{1}{3}{0.480}{-0.020}
\SignedCell{2}{3}{0.480}{-0.020}
\DiagCell{3}{3}
\SignedCell{4}{3}{0.517}{0.017}
\SignedCell{5}{3}{0.621}{0.121}
\SignedCell{6}{3}{0.602}{0.102}

% Row 5: PE
\SignedCell{0}{2}{0.458}{-0.042}
\SignedCell{1}{2}{0.463}{-0.037}
\SignedCell{2}{2}{0.463}{-0.037}
\SignedCell{3}{2}{0.483}{-0.017}
\DiagCell{4}{2}
\SignedCell{5}{2}{0.605}{0.105}
\SignedCell{6}{2}{0.585}{0.085}

% Row 6: P(True)
\SignedCell{0}{1}{0.353}{-0.147}
\SignedCell{1}{1}{0.358}{-0.142}
\SignedCell{2}{1}{0.359}{-0.141}
\SignedCell{3}{1}{0.379}{-0.121}
\SignedCell{4}{1}{0.395}{-0.105}
\DiagCell{5}{1}
\SignedCell{6}{1}{0.481}{-0.019}

% Row 7: ER
\SignedCell{0}{0}{0.373}{-0.127}
\SignedCell{1}{0}{0.378}{-0.122}
\SignedCell{2}{0}{0.378}{-0.122}
\SignedCell{3}{0}{0.398}{-0.102}
\SignedCell{4}{0}{0.415}{-0.085}
\SignedCell{5}{0}{0.519}{0.019}
\DiagCell{6}{0}

\HeatmapLabels

\newcommand{\DeltaLegendColorbar}{%
  \begin{scope}[shift={(7.55,0)}]
    \foreach \k in {0,...,120}{
      \pgfmathsetmacro{\delta}{-0.15 + 0.30*\k/120}
      \pgfmathsetmacro{\colorvalue}{0.5 + \delta}
      \pgfmathsetmacro{\yone}{7*\k/120}
      \pgfmathsetmacro{\ytwo}{7*(\k+1)/120}
      \CellColor{\colorvalue}
      \draw[draw=\cellfill, fill=\cellfill] (0,\yone) rectangle (0.25,\ytwo);
    }
    \draw[black, line width=0.3pt] (0,0) rectangle (0.25,7);

    \node[anchor=west, font=\scriptsize] at (0.35,0.00) {$-0.15$};
    \node[anchor=west, font=\scriptsize] at (0.35,1.17) {$-0.10$};
    \node[anchor=west, font=\scriptsize] at (0.35,2.33) {$-0.05$};
    \node[anchor=west, font=\scriptsize] at (0.35,3.50) {$0.00$};
    \node[anchor=west, font=\scriptsize] at (0.35,4.67) {$0.05$};
    \node[anchor=west, font=\scriptsize] at (0.35,5.83) {$0.10$};
    \node[anchor=west, font=\scriptsize] at (0.35,7.00) {$0.15$};

    \node[rotate=90, font=\scriptsize] at (1.05,3.5) {Avg.\ $\Delta$AUROC};
  \end{scope}
}

\end{tikzpicture}
}
\caption{Pairwise average $\Delta$AUROC across 20 model--dataset settings. Each cell shows the average AUROC difference between the row method and the column method, computed as $\mathrm{AUROC}_{\text{row}}-\mathrm{AUROC}_{\text{column}}$. Positive values indicate that the row method performs better on average. Colors reuse the same red--white--blue scale as the win-rate heatmap, centered at zero average difference.}
\label{fig:pairwise_delta_auroc}
\end{figure}

\section{Limitations}

While the proposed method improves the semantic clustering stage used for entropy-based uncertainty estimation, several limitations remain. The method requires access to token or sequence probabilities to compute the probability-weighted entropy score. Therefore, it is less directly applicable to fully black-box LLMs that only return generated text without likelihood information. Emb-SE avoids pairwise NLI inference during clustering. For n sampled responses, entailment-based clustering requires multiple pairwise comparisons, whereas Emb-SE embeds each response once and performs distance-based clustering in the embedding space. Although K-Medoids adds clustering overhead, the method avoids repeated DeBERTa inference and provides directly inspectable cluster geometry.

\section{Conclusion}

We proposed \textit{Embedding Semantic Entropy}, a new embedding-based semantic uncertainty estimator that replaces entailment-based clustering with geometry-aware clustering in embedding space. The resulting cluster distribution is used to compute an entropy-based uncertainty score. By using K-Medoids clustering and dynamic cluster selection with compactness constraints and elbow-based analysis, the proposed method accounts for both intra-cluster coherence and inter-cluster separation.

Experimental results across multiple LLMs and QA datasets show that the proposed approach achieves competitive performance across AUROC and AUARC. These findings suggest that incorporating embedding-space geometry can improve semantic grouping for uncertainty estimation. Moreover, the results indicate that DeBERTa-based semantic clustering is not strictly necessary for competitive semantic uncertainty estimation, as embedding-based clustering achieves comparable performance while offering a more transparent clustering process.

% ACML/JMLR template handles the bibliography style.
% \bibliographystyle{plainnat}
\bibliography{reference_akshi}

\appendix

\section{Answer Generation Prompt}
\label{app:generation_prompt}

This appendix describes the prompt format used to generate model answers for the uncertainty estimation experiments. For each dataset example, the prompt consists of a brief instruction, optional few-shot demonstrations, and the target question. The brief instruction used in the default setting is:
\begin{quote}
\texttt{Answer the following question as briefly as possible.}
\end{quote}
For few-shot prompting, examples are sampled from the training set. Each few-shot example is formatted as a question--answer pair, optionally including context when the dataset setting requires it. The general few-shot format is:
\begin{quote}
\texttt{Question: <question>}\\
\texttt{Answer: <answer>}
\end{quote}
When context is enabled, the example format becomes:
\begin{quote}
\texttt{Context: <context>}\\
\texttt{Question: <question>}\\
\texttt{Answer: <answer>}
\end{quote}
For the target validation example, the same format is used, but the answer field is left open for the model to complete. Without context, the final query prompt is:
\begin{quote}
\texttt{Question: <question>}\\
\texttt{Answer:}
\end{quote}
With context, the final query prompt is:
\begin{quote}
\texttt{Context: <context>}\\
\texttt{Question: <question>}\\
\texttt{Answer:}
\end{quote}

Thus, the full prompt used for generation is:
\begin{equation}
\text{Prompt}
=
\text{Brief instruction}
+
\text{Few-shot examples}
+
\text{Target question}.
\end{equation}

In the experiments, five few-shot examples are used by default. For each target question, one low-temperature response is generated with temperature $T=0.1$ and used for accuracy evaluation. In addition, ten high-temperature responses are generated with temperature $T=1.0$ and used for semantic uncertainty estimation. The maximum generation length is set to 50 new tokens. This setup follows the uncertainty estimation protocol in which the low-temperature response represents the model's most likely answer, while the high-temperature responses provide a sample of possible semantic alternatives.

\section{Hyperparameter Selection Details}
\label{app:hyperparameter_selection}

Table~\ref{tab:hyperparameter_summary} summarizes the hyperparameter values selected by 5-fold cross-validation. The search space is fixed for all model--dataset settings, with $\tau \in [0.10,0.50]$ and $\gamma \in [0.01,0.50]$. For each model--dataset setting, cross-validation selects the values that maximize validation AUROC.

The table is a compact summary of these model--dataset-specific selections rather than a full listing of every selected pair. The compactness threshold $\tau$ is grouped by model because it mainly reflects model-specific variation in response dispersion within the embedding space. The elbow sensitivity threshold $\gamma$ is grouped by dataset because it reflects how clearly generated answers separate into semantic clusters for each benchmark. Although both parameters may be influenced by both model and dataset, this grouping highlights the dominant variation observed in the selected values.

In Table~\ref{tab:hyperparameter_summary}, the range reports the minimum and maximum values chosen by cross-validation within each group, not the full grid-search range. The avg selected value is computed over all cross-validation-selected hyperparameter values within each group and is therefore not necessarily equal to the midpoint of the selected range. These selected values are used for the final Embedding Semantic Entropy evaluation reported in the main results.

\begin{table}[htbp]
\centering
\caption{Summary of selected hyperparameters after 5-fold cross-validation. The selected range reports the minimum and maximum values chosen within each group, not the full grid-search range.}
\label{tab:hyperparameter_summary}
\small
\setlength{\tabcolsep}{4pt}
\renewcommand{\arraystretch}{1.1}
\begin{tabular}{llcc}
\toprule
\textbf{Parameter} & \textbf{Group} & \textbf{Avg. selected value} & \textbf{Selected range} \\
\midrule
$\gamma$ & NQ & 0.39 & 0.23--0.50 \\
$\gamma$ & SQuAD & 0.34 & 0.23--0.39 \\
$\gamma$ & SVAMP & 0.28 & 0.05--0.46 \\
$\gamma$ & TriviaQA & 0.25 & 0.14--0.44 \\
$\gamma$ & BioASQ & 0.21 & 0.01--0.42 \\
\midrule
$\tau$ & Gemma-7B & 0.16 & 0.15--0.20 \\
$\tau$ & LLaMA-2-7B-Chat & 0.23 & 0.20--0.25 \\
$\tau$ & LLaMA-2-7B & 0.13 & 0.10--0.20 \\
$\tau$ & Falcon-7B-Instruct & 0.11 & 0.10--0.15 \\
\bottomrule
\end{tabular}
\end{table}

\section{Full Experimental Results}
\label{app:full_results}

This appendix provides the complete numerical results corresponding to the AUROC and AUARC bar plots presented in the main text.
Tables~\ref{tab:falcon7b-instruct-auroc-auarc}--\ref{tab:gemma7b-auroc-auarc} report the mean performance and standard deviation for each uncertainty estimation method across all model--dataset settings.
These results include the proposed Embedding Semantic Entropy method together with the baseline methods SE, DSE, KLE, PE, $P(\mathrm{True})$, and ER.
The best-performing method for each model--dataset setting is highlighted in bold.

\vspace{1em}

\begin{table}[t]
\centering
\caption{Falcon-7B-Instruct AUROC and AUARC scores across datasets. Best results per column are shown in bold.}
\label{tab:falcon7b-instruct-auroc-auarc}
\small
\setlength{\tabcolsep}{4pt}
\renewcommand{\arraystretch}{1.1}
\resizebox{\textwidth}{!}{%
\begin{tabular}{@{}l*{10}{c}@{}}
\toprule
\multirow{2}{*}{Method} &
\multicolumn{2}{c}{TriviaQA} &
\multicolumn{2}{c}{SQuAD} &
\multicolumn{2}{c}{NQ} &
\multicolumn{2}{c}{SVAMP} &
\multicolumn{2}{c}{BioASQ} \\
\cmidrule(lr){2-3}
\cmidrule(lr){4-5}
\cmidrule(lr){6-7}
\cmidrule(lr){8-9}
\cmidrule(l){10-11}
& AUROC & AUARC & AUROC & AUARC & AUROC & AUARC & AUROC & AUARC & AUROC & AUARC \\
\midrule
SE & 0.791 $\pm$ 0.02 & \textbf{0.549 $\pm$ 0.03} & 0.665 $\pm$ 0.04 & 0.251 $\pm$ 0.03 & 0.716 $\pm$ 0.03 & 0.369 $\pm$ 0.03 & 0.698 $\pm$ 0.04 & 0.279 $\pm$ 0.04 & 0.831 $\pm$ 0.02 & 0.499 $\pm$ 0.03 \\
DSE & 0.779 $\pm$ 0.02 & 0.532 $\pm$ 0.03 & 0.655 $\pm$ 0.04 & 0.241 $\pm$ 0.03 & 0.714 $\pm$ 0.03 & 0.351 $\pm$ 0.03 & 0.711 $\pm$ 0.04 & 0.267 $\pm$ 0.03 & 0.828 $\pm$ 0.02 & 0.479 $\pm$ 0.03 \\
PE & 0.750 $\pm$ 0.03 & 0.522 $\pm$ 0.03 & 0.665 $\pm$ 0.04 & \textbf{0.257 $\pm$ 0.03} & 0.671 $\pm$ 0.03 & 0.348 $\pm$ 0.03 & \textbf{0.732 $\pm$ 0.04} & \textbf{0.290 $\pm$ 0.03} & 0.655 $\pm$ 0.03 & 0.406 $\pm$ 0.03 \\
$P_{\mathrm{true}}$ & 0.544 $\pm$ 0.03 & 0.390 $\pm$ 0.03 & 0.404 $\pm$ 0.03 & 0.133 $\pm$ 0.02 & 0.554 $\pm$ 0.03 & 0.280 $\pm$ 0.03 & 0.592 $\pm$ 0.04 & 0.226 $\pm$ 0.03 & 0.598 $\pm$ 0.03 & 0.358 $\pm$ 0.03 \\
ER & 0.719 $\pm$ 0.03 & 0.490 $\pm$ 0.03 & 0.628 $\pm$ 0.04 & 0.232 $\pm$ 0.03 & 0.658 $\pm$ 0.03 & 0.325 $\pm$ 0.03 & 0.683 $\pm$ 0.04 & 0.258 $\pm$ 0.03 & 0.829 $\pm$ 0.02 & \textbf{0.507 $\pm$ 0.03} \\
KLE($K_{\mathrm{HEAT}}$) & 0.771 $\pm$ 0.02 & 0.537 $\pm$ 0.03 & 0.602 $\pm$ 0.04 & 0.223 $\pm$ 0.03 & 0.712 $\pm$ 0.03 & 0.362 $\pm$ 0.03 & 0.706 $\pm$ 0.04 & 0.281 $\pm$ 0.03 & 0.785 $\pm$ 0.02 & 0.468 $\pm$ 0.03 \\
Emb-SE & \textbf{0.796 $\pm$ 0.02} & \textbf{0.549 $\pm$ 0.03} & \textbf{0.671 $\pm$ 0.04} & 0.255 $\pm$ 0.03 & \textbf{0.733 $\pm$ 0.03} & \textbf{0.378 $\pm$ 0.03} & 0.694 $\pm$ 0.04 & 0.277 $\pm$ 0.03 & \textbf{0.834 $\pm$ 0.02} & 0.504 $\pm$ 0.03 \\
\bottomrule
\end{tabular}%
}
\end{table}

\begin{table}[t]
\centering
\caption{Llama-2-7B AUROC and AUARC scores across datasets. Best results per column are shown in bold.}
\label{tab:llama2-7b-auroc-auarc}
\small
\setlength{\tabcolsep}{4pt}
\renewcommand{\arraystretch}{1.1}
\resizebox{\textwidth}{!}{%
\begin{tabular}{@{}l*{10}{c}@{}}
\toprule
\multirow{2}{*}{Method} &
\multicolumn{2}{c}{TriviaQA} &
\multicolumn{2}{c}{SQuAD} &
\multicolumn{2}{c}{NQ} &
\multicolumn{2}{c}{SVAMP} &
\multicolumn{2}{c}{BioASQ} \\
\cmidrule(lr){2-3}
\cmidrule(lr){4-5}
\cmidrule(lr){6-7}
\cmidrule(lr){8-9}
\cmidrule(l){10-11}
& AUROC & AUARC & AUROC & AUARC & AUROC & AUARC & AUROC & AUARC & AUROC & AUARC \\
\midrule
SE & 0.784 $\pm$ 0.03 & 0.745 $\pm$ 0.02 & 0.785 $\pm$ 0.03 & 0.405 $\pm$ 0.03 & 0.775 $\pm$ 0.02 & 0.541 $\pm$ 0.03 & \textbf{0.878 $\pm$ 0.02} & \textbf{0.649 $\pm$ 0.03} & 0.852 $\pm$ 0.02 & 0.642 $\pm$ 0.03 \\
DSE & 0.778 $\pm$ 0.02 & 0.742 $\pm$ 0.02 & \textbf{0.798 $\pm$ 0.03} & 0.398 $\pm$ 0.03 & 0.776 $\pm$ 0.02 & 0.527 $\pm$ 0.03 & 0.870 $\pm$ 0.02 & 0.636 $\pm$ 0.03 & 0.848 $\pm$ 0.02 & 0.618 $\pm$ 0.03 \\
PE & 0.751 $\pm$ 0.03 & 0.730 $\pm$ 0.02 & 0.757 $\pm$ 0.03 & \textbf{0.408 $\pm$ 0.03} & 0.750 $\pm$ 0.03 & 0.533 $\pm$ 0.03 & 0.841 $\pm$ 0.02 & 0.632 $\pm$ 0.03 & 0.819 $\pm$ 0.02 & 0.620 $\pm$ 0.03 \\
$P_{\mathrm{true}}$ & 0.607 $\pm$ 0.03 & 0.668 $\pm$ 0.03 & 0.628 $\pm$ 0.03 & 0.331 $\pm$ 0.03 & 0.678 $\pm$ 0.03 & 0.480 $\pm$ 0.03 & 0.751 $\pm$ 0.03 & 0.555 $\pm$ 0.04 & 0.672 $\pm$ 0.03 & 0.521 $\pm$ 0.03 \\
ER & 0.619 $\pm$ 0.03 & 0.664 $\pm$ 0.03 & 0.573 $\pm$ 0.03 & 0.288 $\pm$ 0.03 & 0.569 $\pm$ 0.03 & 0.403 $\pm$ 0.03 & 0.758 $\pm$ 0.03 & 0.574 $\pm$ 0.03 & 0.748 $\pm$ 0.02 & 0.583 $\pm$ 0.03 \\
KLE($K_{\mathrm{HEAT}}$) & 0.775 $\pm$ 0.03 & 0.743 $\pm$ 0.02 & 0.710 $\pm$ 0.03 & 0.381 $\pm$ 0.03 & 0.780 $\pm$ 0.02 & 0.542 $\pm$ 0.03 & 0.867 $\pm$ 0.02 & 0.642 $\pm$ 0.03 & 0.818 $\pm$ 0.02 & 0.623 $\pm$ 0.03 \\
Emb-SE & \textbf{0.801 $\pm$ 0.02} & \textbf{0.755 $\pm$ 0.02} & 0.778 $\pm$ 0.03 & 0.403 $\pm$ 0.03 & \textbf{0.785 $\pm$ 0.02} & \textbf{0.544 $\pm$ 0.03} & 0.868 $\pm$ 0.02 & 0.644 $\pm$ 0.03 & \textbf{0.858 $\pm$ 0.02} & \textbf{0.644 $\pm$ 0.03} \\
\bottomrule
\end{tabular}%
}
\end{table}

\begin{table}[t]
\centering
\caption{Llama-2-7B-Chat AUROC and AUARC scores across datasets. Best results per column are shown in bold.}
\label{tab:llama2-7b-chat-auroc-auarc}
\small
\setlength{\tabcolsep}{4pt}
\renewcommand{\arraystretch}{1.1}
\resizebox{\textwidth}{!}{%
\begin{tabular}{@{}l*{10}{c}@{}}
\toprule
\multirow{2}{*}{Method} &
\multicolumn{2}{c}{TriviaQA} &
\multicolumn{2}{c}{SQuAD} &
\multicolumn{2}{c}{NQ} &
\multicolumn{2}{c}{SVAMP} &
\multicolumn{2}{c}{BioASQ} \\
\cmidrule(lr){2-3}
\cmidrule(lr){4-5}
\cmidrule(lr){6-7}
\cmidrule(lr){8-9}
\cmidrule(l){10-11}
& AUROC & AUARC & AUROC & AUARC & AUROC & AUARC & AUROC & AUARC & AUROC & AUARC \\
\midrule
SE & 0.787 $\pm$ 0.02 & 0.731 $\pm$ 0.02 & 0.750 $\pm$ 0.03 & 0.360 $\pm$ 0.03 & 0.724 $\pm$ 0.03 & \textbf{0.514 $\pm$ 0.03} & 0.841 $\pm$ 0.02 & 0.590 $\pm$ 0.03 & 0.842 $\pm$ 0.02 & 0.640 $\pm$ 0.03 \\
DSE & \textbf{0.788 $\pm$ 0.02} & 0.730 $\pm$ 0.02 & 0.755 $\pm$ 0.03 & 0.356 $\pm$ 0.03 & 0.726 $\pm$ 0.03 & 0.510 $\pm$ 0.03 & \textbf{0.845 $\pm$ 0.02} & 0.588 $\pm$ 0.03 & 0.840 $\pm$ 0.02 & 0.632 $\pm$ 0.03 \\
PE & 0.758 $\pm$ 0.02 & 0.722 $\pm$ 0.02 & 0.719 $\pm$ 0.03 & 0.359 $\pm$ 0.03 & 0.715 $\pm$ 0.03 & 0.510 $\pm$ 0.03 & 0.835 $\pm$ 0.02 & 0.589 $\pm$ 0.03 & 0.678 $\pm$ 0.03 & 0.554 $\pm$ 0.03 \\
$P_{\mathrm{true}}$ & 0.670 $\pm$ 0.03 & 0.662 $\pm$ 0.03 & 0.598 $\pm$ 0.04 & 0.297 $\pm$ 0.03 & 0.647 $\pm$ 0.03 & 0.465 $\pm$ 0.03 & 0.573 $\pm$ 0.03 & 0.422 $\pm$ 0.04 & 0.790 $\pm$ 0.02 & 0.633 $\pm$ 0.03 \\
ER & 0.599 $\pm$ 0.02 & 0.636 $\pm$ 0.03 & 0.562 $\pm$ 0.03 & 0.266 $\pm$ 0.03 & 0.570 $\pm$ 0.03 & 0.412 $\pm$ 0.03 & 0.803 $\pm$ 0.03 & 0.575 $\pm$ 0.03 & 0.729 $\pm$ 0.03 & 0.569 $\pm$ 0.03 \\
KLE($K_{\mathrm{HEAT}}$) & 0.783 $\pm$ 0.02 & 0.732 $\pm$ 0.02 & 0.707 $\pm$ 0.03 & 0.349 $\pm$ 0.03 & 0.715 $\pm$ 0.03 & 0.508 $\pm$ 0.03 & \textbf{0.845 $\pm$ 0.02} & \textbf{0.591 $\pm$ 0.03} & 0.805 $\pm$ 0.02 & 0.623 $\pm$ 0.03 \\
Emb-SE & \textbf{0.788 $\pm$ 0.02} & \textbf{0.733 $\pm$ 0.02} & \textbf{0.762 $\pm$ 0.03} & \textbf{0.368 $\pm$ 0.03} & \textbf{0.728 $\pm$ 0.03} & 0.513 $\pm$ 0.03 & 0.827 $\pm$ 0.02 & 0.579 $\pm$ 0.03 & \textbf{0.845 $\pm$ 0.02} & \textbf{0.646 $\pm$ 0.03} \\
\bottomrule
\end{tabular}%
}
\end{table}

\begin{table}[t]
\centering
\caption{Gemma-7B AUROC and AUARC scores across datasets. Best results per column are shown in bold.}
\label{tab:gemma7b-auroc-auarc}
\small
\setlength{\tabcolsep}{4pt}
\renewcommand{\arraystretch}{1.1}
\resizebox{\textwidth}{!}{%
\begin{tabular}{@{}l*{10}{c}@{}}
\toprule
\multirow{2}{*}{Method} &
\multicolumn{2}{c}{TriviaQA} &
\multicolumn{2}{c}{SQuAD} &
\multicolumn{2}{c}{NQ} &
\multicolumn{2}{c}{SVAMP} &
\multicolumn{2}{c}{BioASQ} \\
\cmidrule(lr){2-3}
\cmidrule(lr){4-5}
\cmidrule(lr){6-7}
\cmidrule(lr){8-9}
\cmidrule(l){10-11}
& AUROC & AUARC & AUROC & AUARC & AUROC & AUARC & AUROC & AUARC & AUROC & AUARC \\
\midrule
SE & 0.767 $\pm$ 0.02 & 0.705 $\pm$ 0.03 & 0.744 $\pm$ 0.03 & 0.364 $\pm$ 0.03 & 0.735 $\pm$ 0.03 & 0.528 $\pm$ 0.03 & 0.890 $\pm$ 0.02 & 0.763 $\pm$ 0.02 & 0.869 $\pm$ 0.02 & \textbf{0.702 $\pm$ 0.02} \\
DSE & 0.764 $\pm$ 0.02 & 0.695 $\pm$ 0.02 & 0.749 $\pm$ 0.03 & 0.350 $\pm$ 0.03 & \textbf{0.750 $\pm$ 0.03} & 0.522 $\pm$ 0.03 & 0.878 $\pm$ 0.02 & 0.749 $\pm$ 0.03 & 0.865 $\pm$ 0.02 & 0.683 $\pm$ 0.02 \\
PE & 0.744 $\pm$ 0.02 & 0.698 $\pm$ 0.02 & 0.737 $\pm$ 0.03 & 0.361 $\pm$ 0.03 & 0.725 $\pm$ 0.03 & 0.513 $\pm$ 0.03 & 0.858 $\pm$ 0.02 & 0.753 $\pm$ 0.03 & 0.825 $\pm$ 0.02 & 0.686 $\pm$ 0.02 \\
$P_{\mathrm{true}}$ & 0.739 $\pm$ 0.03 & 0.690 $\pm$ 0.03 & 0.644 $\pm$ 0.03 & 0.328 $\pm$ 0.03 & 0.684 $\pm$ 0.03 & 0.501 $\pm$ 0.03 & 0.710 $\pm$ 0.03 & 0.649 $\pm$ 0.03 & 0.808 $\pm$ 0.02 & 0.681 $\pm$ 0.02 \\
ER & 0.647 $\pm$ 0.02 & 0.625 $\pm$ 0.03 & 0.575 $\pm$ 0.03 & 0.298 $\pm$ 0.03 & 0.598 $\pm$ 0.03 & 0.416 $\pm$ 0.03 & 0.627 $\pm$ 0.02 & 0.604 $\pm$ 0.03 & 0.784 $\pm$ 0.02 & 0.547 $\pm$ 0.02 \\
KLE($K_{\mathrm{HEAT}}$) & 0.762 $\pm$ 0.02 & 0.701 $\pm$ 0.02 & 0.749 $\pm$ 0.03 & 0.340 $\pm$ 0.03 & 0.739 $\pm$ 0.03 & 0.526 $\pm$ 0.03 & 0.884 $\pm$ 0.02 & 0.760 $\pm$ 0.02 & 0.804 $\pm$ 0.02 & 0.670 $\pm$ 0.02 \\
Emb-SE & \textbf{0.781 $\pm$ 0.02} & \textbf{0.716 $\pm$ 0.02} & \textbf{0.767 $\pm$ 0.03} & \textbf{0.378 $\pm$ 0.03} & 0.745 $\pm$ 0.03 & \textbf{0.539 $\pm$ 0.03} & \textbf{0.891 $\pm$ 0.02} & \textbf{0.765 $\pm$ 0.02} & \textbf{0.875 $\pm$ 0.02} & \textbf{0.702 $\pm$ 0.02} \\
\bottomrule
\end{tabular}%
}
\end{table}

\end{document}